\documentclass[conference]{IEEEtran}
\IEEEoverridecommandlockouts
\usepackage{cite}
\usepackage{amsmath,amssymb,amsfonts}
\usepackage{algorithmic}
\usepackage{graphicx}
\usepackage{textcomp}
\usepackage{xcolor}

\usepackage{graphics} 
\usepackage{caption}

\usepackage{svg}
\usepackage{float}
\usepackage{subfigure}
\usepackage{comment}
\usepackage[inline]{enumitem}
\usepackage{hyperref}
\usepackage{etoolbox}
\usepackage{wrapfig}

\def\BibTeX{{\rm B\kern-.05em{\sc i\kern-.025em b}\kern-.08em
    T\kern-.1667em\lower.7ex\hbox{E}\kern-.125emX}}

\begin{document}

\title{
{\bf COR-MP}: {\bf C}onservation {\bf o}f {\bf R}esources Model for {\bf M}aneuver {\bf P}lanning}

\author{\IEEEauthorblockN{Karim Essalmi}
\IEEEauthorblockA{\textit{ASTRA Team, Inria \& Valeo} \\
Paris \& Créteil, France \\
karim.essalmi@inria.fr \\
karim.essalmi@valeo.com}
\and
\IEEEauthorblockN{Fernando Garrido}
\IEEEauthorblockA{\textit{ASTRA Team, Valeo} \\
Créteil, France \\
fernando.garrido@valeo.com}
\and
\IEEEauthorblockN{Fawzi Nashashibi}
\IEEEauthorblockA{\textit{ASTRA Team, Inria} \\
Paris, France \\
fawzi.nashashibi@inria.fr}
}

\maketitle

\begin{abstract}
Decision-making for automated driving remains a challenging task. For their integration into real platforms, these algorithms must guarantee passenger safety and comfort while ensuring interpretability and an appropriate computational time. To model and solve this decision-making problem, we have developed a novel approach called COR-MP (Conservation of Resources model for Maneuver Planning). This model is based on the Conservation of Resources theory, a psychological concept applied to human behavior. COR-MP is based on various driving parameters, such as comfort, safety, or energy, and provides in real-time a profit value that enables us to quantify the impact of a decision on the decision-maker. Our method has been tested and validated through closed-loop simulations using RTMaps middleware, and preliminary results have been obtained by testing COR-MP on a real vehicle. 

\end{abstract}


\section{Introduction}

Deployment of fully automated vehicles might be viewed as futuristic, but with the recent progress in the domain, it is becoming increasingly achievable. It continues to provoke considerable interest among companies and researchers \cite{HARB}. Indeed, the integration of self-driving cars in our daily lives can bring several benefits, such as a potential reduction of accidents \cite{dixon2020drives}, since a large number of them are due to human error \cite{StatsAccident}, enhance the quality and efficiency of time spent in cars \cite{ReviewDM}, and an improvement in transportation accessibility \cite{smith2015benefits}.  

Despite recent advancements in the domain, there are still challenges to reach levels 4 (highly automated) and 5 (fully automated) of automation notably in the field of decision-making. Modeling human behavior through a machine and finding an optimal decision depending on a large scale of parameters are still challenging \cite{GapDM}.

Decision-making can be seen as the brain of the automated vehicle system \cite{wang2024survey}. When discussing decision-making in the context of self-driving vehicles, it typically refers to four components \cite{Fernando, DecisionMakingClassification}: (a) \textit{Route Planning}: to generate an efficient and effective path from a starting point to the desired destination; (b) \textit{Maneuver Planning}: to establish optimal and safe sequences of vehicle movements; (c) \textit{Trajectory Planning}: generate a feasible and optimized trajectory while taking into account dynamic constraints and environmental factors; and (d) \textit{Control}: compute the control law for the actuators to track a reference trajectory, path and/or speed profile. 

This study focuses on the tactical part of the decision-making classification. This aspect is in charge of identifying and ranking the possible sequence of maneuvers to be performed by the vehicle. These maneuvers can be: keeping the same lane, overtaking a car, or turning at an intersection. Through the adaptation of the Conservation of Resources (COR) theory \cite{hobfoll1989}, which is a psychological framework, into a computational model. COR-MP is capable of real-time ranking of various feasible maneuvers, thereby assuming the role of a maneuver planner in automated driving. 

The Conservation of Resources theory formalizes the decision-making process of humans when faced to stress. Based on personal resources and their evolution, this theory can predict if a human will face stressful situations or, oppositely well-being behaviors. We find this theory suitable for an application to a behavior planner in autonomous driving and a description of our approach (COR-MP) will be detailed in this study.

As described later in this study, our approach shares similarities with utility-based techniques. However, it differs primarily in two ways: first, in its adaptability to various driving scenarios without requiring extensive tuning; and second, in its interpretability, as each decision made by COR-MP can be explained by analyzing the resources. Our approach can be classified as a psychological human model which can be a sub-model within utility-based techniques. 

The main contributions of this paper are:
\begin{itemize}
    \item A new design to model and solve a decision-making problem for automated driving through the adaptation of the Conservation of Resources theory.
    \item A demonstration of the results obtained using COR-MP as the behavioral layer in a real car illustrates similarities between COR-MP decisions and human decisions.
\end{itemize}

This paper is structured in the following manner: Section \ref{section:RelatedWork} provides a brief overview of previous work and recent advances in tactical decision-making for driverless vehicles. It is followed by an introduction to COR theory and its main field of application in Section \ref{section:COR}. Our novel approach is then presented in Section \ref{section:COR-MP}, with Section \ref{section:Results} containing some results obtained in simulation and others obtained by implementing COR-MP into a real vehicle on highway driving. Finally, conclusions and directions for future research are discussed in Section \ref{section:Conclu&FutureWork}.

\section{Related Work}
\label{section:RelatedWork}
In the field of behavior planning, various approaches have been proposed and documented in the literature. These different methods can be classified into two categories \cite{ManeuverPlannerReview}: \textit{Classical} and \textit{Learning-based}. The first category typically relies on predefined rules (miming a certain behavior) or algorithms based on heuristics to evaluate which decision has to be made by the ego-vehicle (EV). Meanwhile, the second category involves the use of machine learning techniques, such as neural networks, to analyze the situation and adapt the decision based on a learned policy. Methods based on learning need a training stage to be functional and they often suffer from a lack of interpretability \cite{ghoul2023interpretable}.

\begin{itemize}

    \item \textbf{Classical approaches}: 
    
    \begin{enumerate}[label=\alph*)]
        \item \textbf{Rule-based}:
        As mentioned in \cite{ReviewDM} most of the existing behavior planners are based on Finite State Machine (FSM). The environment is usually modeled using a finite number of states and transitions. The transitions between these states depend on the system inputs and the pre-defined conditions. In \cite{rulebased}, authors implemented a maneuver planner based on a hierarchical state machine handling traffic rules and different driving scenarios such as turns, intersections, or passing. They applied their algorithms to the 2007 DARPA Urban Challenge \cite{DarpaUrban}. Rule-based approaches are limited when dealing with unknown situations and uncertainty. Furthermore, this kind of technique suffers from a lack of adaptability to various driving scenarios.

        \item \textbf{Optimization-based}:
        These methods can be split into two sub-categories: \textit{utility}, where a cost function evaluates each maneuver based on different criteria. The objective is to optimize this cost function.         
        For example, in \cite{menendez2018courtesy}, authors managed a merge maneuver while cooperating with another vehicle. They use a utility function based on safety and comfort. They also take into account the cooperativeness with another road driver in the utility function. The objective of the ego is to maximize this cost function.
        
        The second is based on \textit{game theory}, where drivers are considered as players and maneuvers as strategies. The goal is to find the strategies that optimize players' utility. Depending on the type of game, it exists different approaches to solve this kind of algorithm. For example, in a non-cooperative game involving two or more players, the optimal solution is to find the Nash Equilibrium.      
        A game-theory approach is implemented in \cite{GameTheoryAnneSpalanzani}, handling the merging maneuver in high-density traffic scenarios. 
        

        \item \textbf{Probabilistic-based}:            
        These techniques are often modeled as a Markov Decision Process (MDP) \cite{nelson}. This allows consideration of uncertainty during the decision-making process. This uncertainty can be aleatoric (due to inherently random effects) and/or epistemic (caused by a lack of knowledge).
        In \cite{POMDP}, authors state the problem as a Partially Observable Markov Decision Process (POMDP) and solve it by combining Monte Carlo Tree Search (MCTS) with an A-star algorithm. The main advantage of using this technique is its ability to handle uncertainty in the decision-making process; however, it is time-consuming due to the computational complexity involved in its implementation.

    \end{enumerate}
    
    \item \textbf{Learning approaches}:
    
    \begin{enumerate}[label=\alph*)]
        \item \textbf{Reinforcement Learning}:
        Reinforcement Learning (RL) technique is a subset of machine learning approaches where an agent learns how to act in the environment through the use of rewards or penalties. This approach needs a training phase where the agent learns from its mistakes and experiences. 
        In \cite{RL}, authors implemented a maneuver planner based on RL. The objective was to incorporate an altruistic parameter into the decision-making process to facilitate cooperation with other interacting vehicles. This was demonstrated on a highway merging ramp by determining whether the ego-vehicle allowed another vehicle to merge by changing lanes or not. The reward is based on Social Value Orientation (SVO) \cite{SVO}, which quantifies the degree of an agent’s selfishness or altruism.
    
        \item \textbf{Deep Learning}:
        Deep Learning is a subset of machine learning that involves training neural networks with multiple layers to learn from datasets. In \cite{IL}, authors use an imitation learning technique, through transformers, to manage overtaking maneuvers in a highway environment. 



    \end{enumerate}
\end{itemize}

Although learning techniques highly depend on extensive data and suffer from the black box problem, they are promising direction in decision-making for automated driving. As mentioned in \cite{trend}, machine learning is the most inspired direction for decision-making in this field.

\section{Conservation of Resources Theory}
\label{section:COR}
Self-driving cars will share roads with human drivers, requiring cooperation between them. An understanding of how humans make decisions while driving is primordial. The Conservation of Resources theory \cite{hobfoll1989} provides a psychological framework for understanding and modeling human behavior, particularly in stressful situations. 

This theory states that each human being possesses a personal pool of resources, and the evolution of this pool of resources will lead to a certain behavior (stress or well-being). A resource is defined as anything perceived as valuable by an individual. Hobfoll divides resources into four kinds: materials (e.g. home, mansion), personal characteristics (e.g. personal orientation, personal traits, personal skills), conditions (e.g. marriage, tenure, seniority), and energies (e.g. time, money, knowledge) \cite{hobfoll1989}. 

The core concept of this theory is that an actual loss of resources induces a stress state for an individual. Oppositely, protecting and gaining new resources will lead to a well-being behavior. So, to avoid stress, an individual has to protect his resources and try to gain new ones. 

COR theory relies on two principles:
\begin{itemize}
    \item \textbf{Primacy of Loss}: It states that losing a resource has a greater impact in terms of stress than winning the same resource. This idea aligns with the prospect theory \cite{prospecttheory}, a behavioral economics theory, which is based on a similar principle. Both theories stem from the loss aversion phenomenon \cite{lossaversion}, which suggests that people tend to strongly prefer avoiding losses over acquiring gains. 
    
    \item \textbf{Resource Investment}: To protect resources from losses and acquire new ones, individuals must invest their resources. For instance, some allocate funds and time resources towards education with the intention of gaining knowledge in the future.

\end{itemize}

Most of the Conservation of Resources literature is related to organizational and psychological contexts \cite{hobfoll2018conservation}. They mainly study how organizational areas impact the psychological aspects of humans in terms of stress and well-being. 

Among all these studies, only one is a computational model \cite{Sabrina}. The authors developed the Conservation of Resources Engine (COR-E) model to simulate agents' behavior in a simulated environment. This is the first work that has demonstrated the possibility of using COR as a computational model. 

According to the literature, we found COR relevant for an application of a decision-making algorithm for self-driving cars. 

\section{Method}
\label{section:COR-MP}

Our model is defined as a 5-tuple \(\Bigl \langle \mathcal{A}, \mathcal{T}, \mathcal{S}, w, \mu \Bigr \rangle\), where \(\mathcal{A}\), \(\mathcal{T}\), and \(\mathcal{S}\) represent respectively, the set of actions considered, the set of resources type, and the set of resource state. \(w\) and \(\mu\) represent the weight and the value of a given resource.

\subsection{Maneuvers}
As previously mentioned, we are interested in maneuver planning for automated driving. To handle both longitudinal and lateral movements, we define the ego action-space as a set of discrete actions \(a_i \in \mathcal{A}\), where \(i=\{1,..., m\}\), with \(m\) the number of actions considered. 

\begin{equation}
    \label{eq:Maneuvers}
    a_i \in \mathcal{A} =     
    \left\{
    \begin{array}{c}
        \text{Change\ Lane\ Left} \\
        \text{Change\ Lane\ Right} \\
        \text{Keep\ Lane\ Accelerate} \\
        \text{Keep\ Lane\ Same\ Speed} \\
        \text{Keep\ Lane\ Decelerate} \\
        \text{Stop}
    \end{array}
    \right\}
\end{equation}

\subsection{Resources}
In our approach, the maneuver assessment is made through resources. They were selected to closely align with the factors that humans consider when making driving decisions. 

\begin{itemize}

    \item We define \(n\) the number of resources as six, and we denote \(t_j \in \mathcal{T}\) the type of a resource, where \(j=\{1,..., n\}\).

    \begin{equation}
    \label{eq:ResourcesSet}
        t_j \in \mathcal{T} = \left\{ 
        \begin{array}{c}
            \text{Safety} \\
            \text{Comfort} \\
            \text{Objective} \\
            \text{A-priori Lane} \\
            \text{Energy} \\
            \text{Crowdedness}
        \end{array} 
        \right\}
    \end{equation}

    \item A \textbf{weight} \(w \in [0,1]\) is a parameter that describes the importance of a resource in the decision-making process relative to others. For example, in a driving scenario, the resource \textit{Safety} will have a higher impact on the final decision compared to the resource \textit{Energy}. This value is quantified using the Rank Order Centroid (ROC) technique \cite{ROC}:
    
    \begin{equation}
    \label{eq:RankOrderCentroid}
    w = \frac{1}{n}\sum_{j=k}^{n} \frac{1}{j}
    \end{equation}
    
    Where \(k\) is the rank of a resource.  We aimed to make COR-MP adaptable to diverse driving situations without necessitating additional parameter tuning. We also wanted COR-MP to accommodate various driver profiles. Table \ref{tab:ResourcesRanking} shows the possible ranking of the resources depending on diverse driver profiles. 

\begin{table}[!h]
\centering
\begin{tabular}{lcccc}
    \textbf{Resource Type} \(t_j\) &\textbf{Regular}&\textbf{Aggressive}&\textbf{Fuel-efficient}\\
    \hline
    Safety&1st&2nd&2nd\\ 
    Comfort&2nd&3rd&3rd\\ 
    Objective&3rd&1st&5th\\ 
    A-priori Lane&4th&4th&4th\\ 
    Energy&5th&5th&1st\\
    Crowdedness&6th&6th&6th\\
    \hline 
\end{tabular}
\caption{Possible resources ranking depending on the driver profile.}
\label{tab:ResourcesRanking}
\end{table}

    \item A \textbf{value} \(\mu \in [0,1]\) corresponds to the benefit that the evaluate action \(a_i\) will generate. A higher value is favored over a lower one, which aligns with the Primacy of Loss principle in COR theory. Figure \ref{fig:muValues} shows the different values for each resource. These values have been tuned to make COR-MP adaptable to diverse driving situations. 

    \begin{itemize}
        \item \textbf{Safety}: This resource describes the danger of a maneuver. Its value is based on safe longitudinal and lateral distances between the ego-vehicle and other interacting road users. A \(\mu\) value is deduced from these two parameters as shown in Figure \ref{fig:muValues} (a).

        \item \textbf{Comfort}: Passenger comfort is an important factor while driving. We quantify this resource parameter by analyzing longitudinal and lateral accelerations. In \cite{Bae_AccelerationProfile}, a study has been conducted resulting in an Occupant's Preference Metric (OPM). Values presented in \ref{fig:muValues} Figure (b) are based on this study. 

        \item \textbf{Objective}: To achieve the mission, we defined a resource \textit{Objective} which corresponds to the distance the ego vehicle will travel. This value is derived from the distance traveled by each maneuver, taking into account the road's speed limit.

        \item \textbf{A-priori lane}: We define an a-priori lane \(L_{ap}\) as the lane that the EV must follow. This value is determined using the lateral difference between the center of the a-priori lane and the center of the lane that the ego will reach by executing the evaluated maneuver. The further the EV deviates from \(L_{ap}\), the lower \(\mu\) will be. 

        \item \textbf{Energy}: Each maneuver results in a certain energy consumption. One of the goals of COR-MP is to be fuel-efficient. This value is computed through the kinetic energy formula:         
        \begin{equation}
            \label{eq:kineticEnergy}
            \Delta E_c=\frac{1}{2}*m*(v_b-v_a)^2
        \end{equation}

        Where \(\Delta E_c\), \(m\), \(v_b\), and \(v_a\) represent respectively the kinetic energy (kJ) generated by the evaluated maneuver, the mass of the vehicle (kg), the speed (m/s) reached at the end of the maneuver, and the speed at the beginning of the maneuver. We assume that a decelerating maneuver does not consume any kinetic energy.  

        \item \textbf{Crowdedness}: To avoid uncertainties caused by other road users (ORU), we define a resource \textit{Crowdedness} which quantifies the density of ORUs that a maneuver will interact with. We prioritize maneuvers that interact with a lower density of users over those with a higher density.

        \begin{figure}[h]
        \subfigure[Safety]{\includegraphics[width=.24\textwidth]{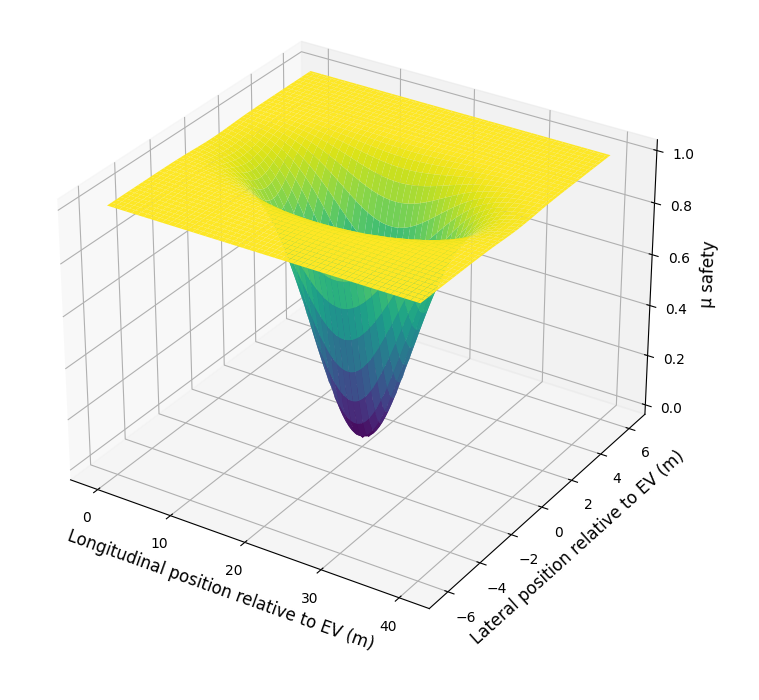}}
        \subfigure[Comfort]{\includegraphics[width=.24\textwidth]{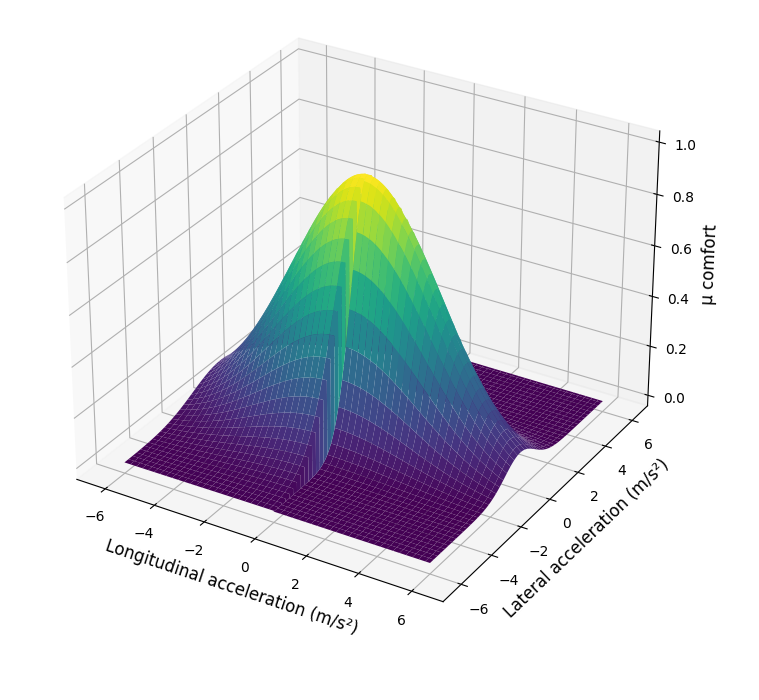}}        
        \subfigure[Objective]{\includegraphics[width=.24\textwidth]{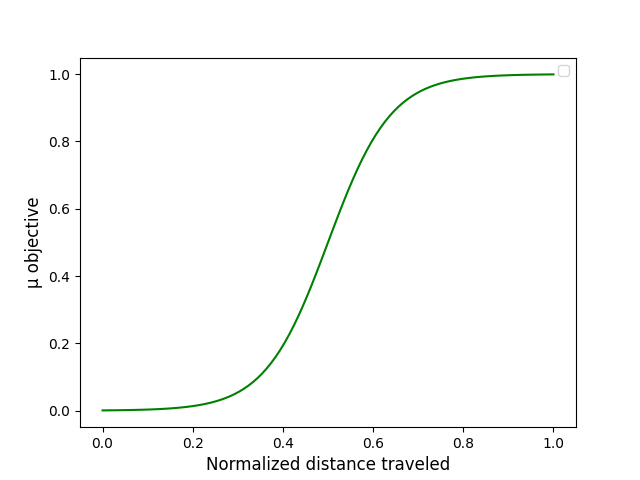}}
        \subfigure[A-priori lane]{\includegraphics[width=.24\textwidth]{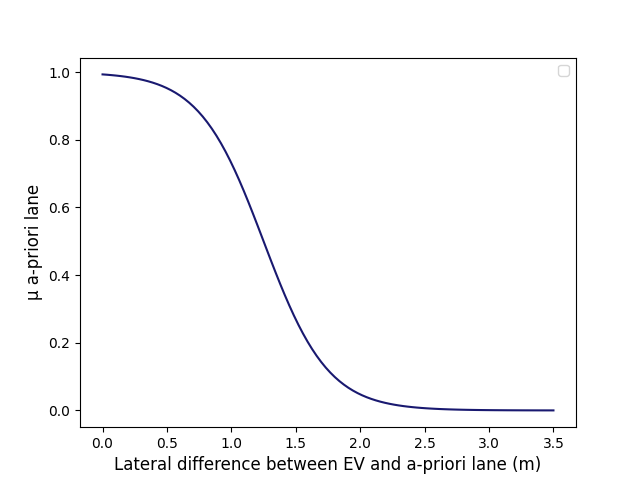}}\\
        \subfigure[Energy]{\includegraphics[width=.24\textwidth]{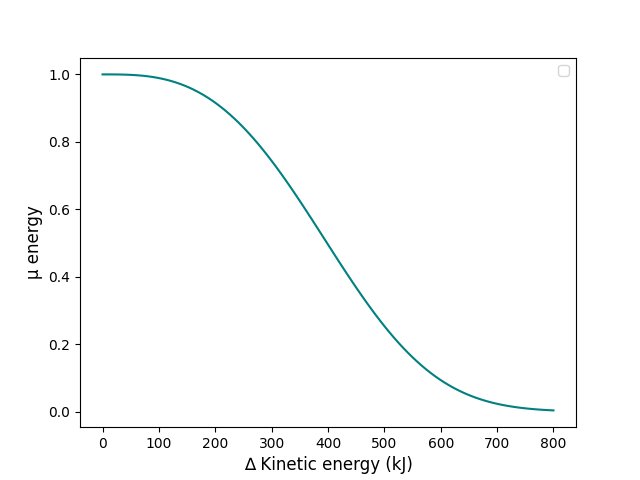}}
        \subfigure[Crowdedness]{\includegraphics[width=.24\textwidth]{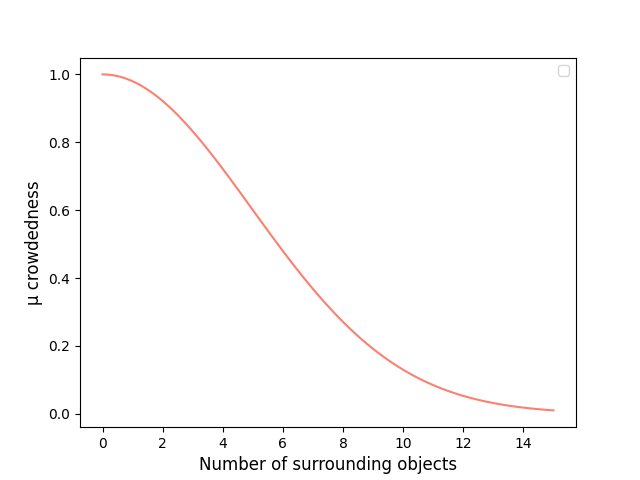}}
        \caption{List of the resources used in the decision evaluation process and their corresponding \(\mu \) values.}
        \label{fig:muValues}
        \end{figure}
    \end{itemize}

    \item A \textbf{state} \(s_l \in \mathcal{S}\), where \(l=\{1,...,4\}\), represents the state of a resource. It allows to describe the evolution of the set of resources \(\mathcal{R}\) by adopting the evaluated action \(a_i\). 
    
    \begin{equation}
    \label{eq:State}
    s_l \in \mathcal{S} =  
    \left\{
    \begin{array}{c}
        \text{Desired}\\
        \text{Acquired} \\
        \text{Threatened} \\
        \text{Loss} \\
    \end{array}
    \right\}
\end{equation}
    
    \(s_l\) is deduced from the resource value \(\mu\). A high resource value indicates the acquisition of the resource, whereas a low value implies a threat or, in the worst case, a loss of the resource. The resource state ensures interpretability, as analyzing \(s_l\) allows us to deduce why a decision has been taken.
        
    

\end{itemize}

\subsection{Methodology}

\begin{figure*}[ht]
\centering
\includegraphics[width=\textwidth]{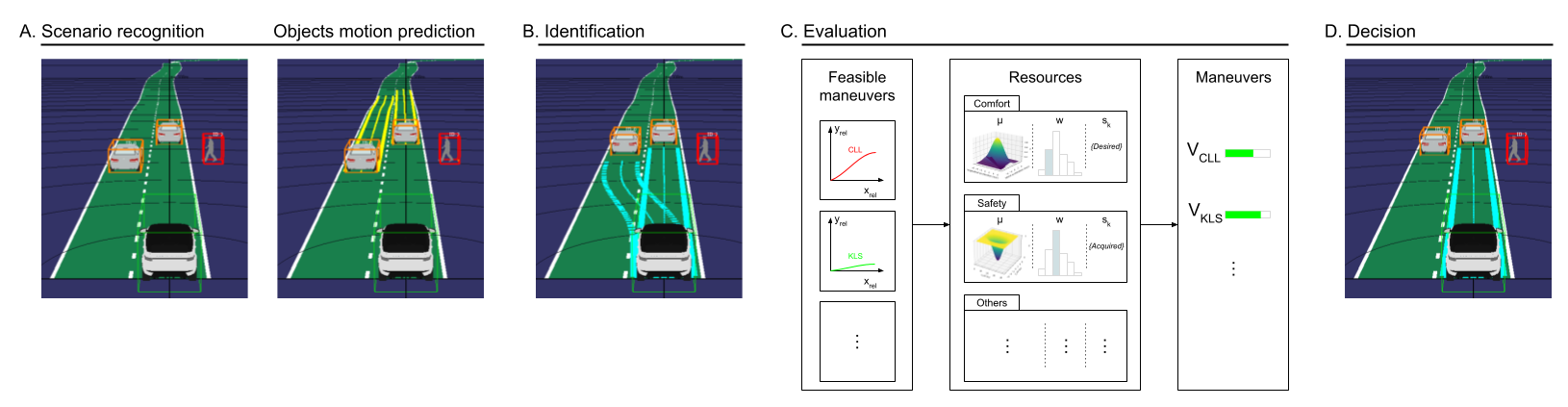}
\caption{Methodology of our approach.}
\label{fig:COR-MP_methodology}
\end{figure*}

As illustrated by Figure \ref{fig:COR-MP_methodology}, COR-MP is composed of four different phases:

\begin{itemize}
    \item \textbf{Scenario Recognition}:
    The goal of this step is twofold: first, to identify both static and dynamic objects that will interact with the EV during the planning horizon. Then, based on this identification, a prediction of their future motion is made. This prediction is based on Cubic Bézier curves under the assumption of constant velocity. The prediction horizon varies depending on the length of the lane and the ego-predicted horizon. The a-priori lane \(L_{ap}\) is also deduced during this stage. 

    \item \textbf{Identification}:
    This step aims to predict the most plausible trajectory that the EV will follow if it decides to execute the corresponding action \(a_i\). The trajectory predicted for the ego is based on Cubic Béziers curves while adapting the different control points depending on the interacting objects. Accuracy in this step is crucial, as these trajectories will be evaluated later. Therefore, they must closely approximate the different trajectories the ego would perform. 

    Computational time is an essential parameter to consider in decision-making algorithms. As mentioned in \cite{Fernando}, at the tactical level the planning horizon is often between 3 and 5 seconds, necessitating a lightweight algorithm. To reduce computational time, a maneuver filter is applied before evaluating ego-actions. 
    Therefore, we characterize a maneuver as \textit{feasible} only if it is collision-free and if it complies with traffic rules, such as traffic lights, traffic signs, and speed limits. The collision-free parameter is determined using the Time To Collision (TTC) metric \cite{metrics}, which assesses the risk of a maneuver considering surrounding objects. The smaller the TTC, the riskier the maneuver.  

    \item \textbf{Evaluation}:
    For each maneuver considered as \textit{feasible}, an assessment through resources is conducted, resulting in the assignment of a profit value \(V_{a_i}\) to each feasible ego-action. 

    \begin{equation}
    V_{a_i}(w,\mu) = \sum_{j=1}^{n}w_j*\mu_j
    \end{equation}

    \item \textbf{Decision}:
    The expected utility theory \cite{expectedUtility} states that rational agents make decisions based on maximizing their personal utility. According to these findings, we model our system as an agent seeking to maximize its profit value. The maneuver with the highest profit value \(a^*\) is then evaluated as the optimal decision to be made by the ego, according to COR-MP. This complies with the first principle of COR Theory, as a high \(\mu\) value implies the acquisition of the resource and thus protects it. 

    \begin{equation}
    \label{eq:Max}
        a^* = \underset{a_i \in \mathcal{A}}{\max} \ V_{a_i}(w, \mu)
    \end{equation}

    COR-MP controls both the speed and the lane selection of the ego-vehicle, influencing its longitudinal and lateral movements. 
    
\end{itemize}

\section{Results}
\label{section:Results}
COR-MP has been tested in simulation in both urban and highway environments and in an open-loop configuration in a real vehicle. This section aims to present some simulation results, as well as those obtained while testing our approach in a real car.

\subsection{Simulation Results}
\begin{figure*}[ht]
\centering
\includegraphics[width=0.8\textwidth]{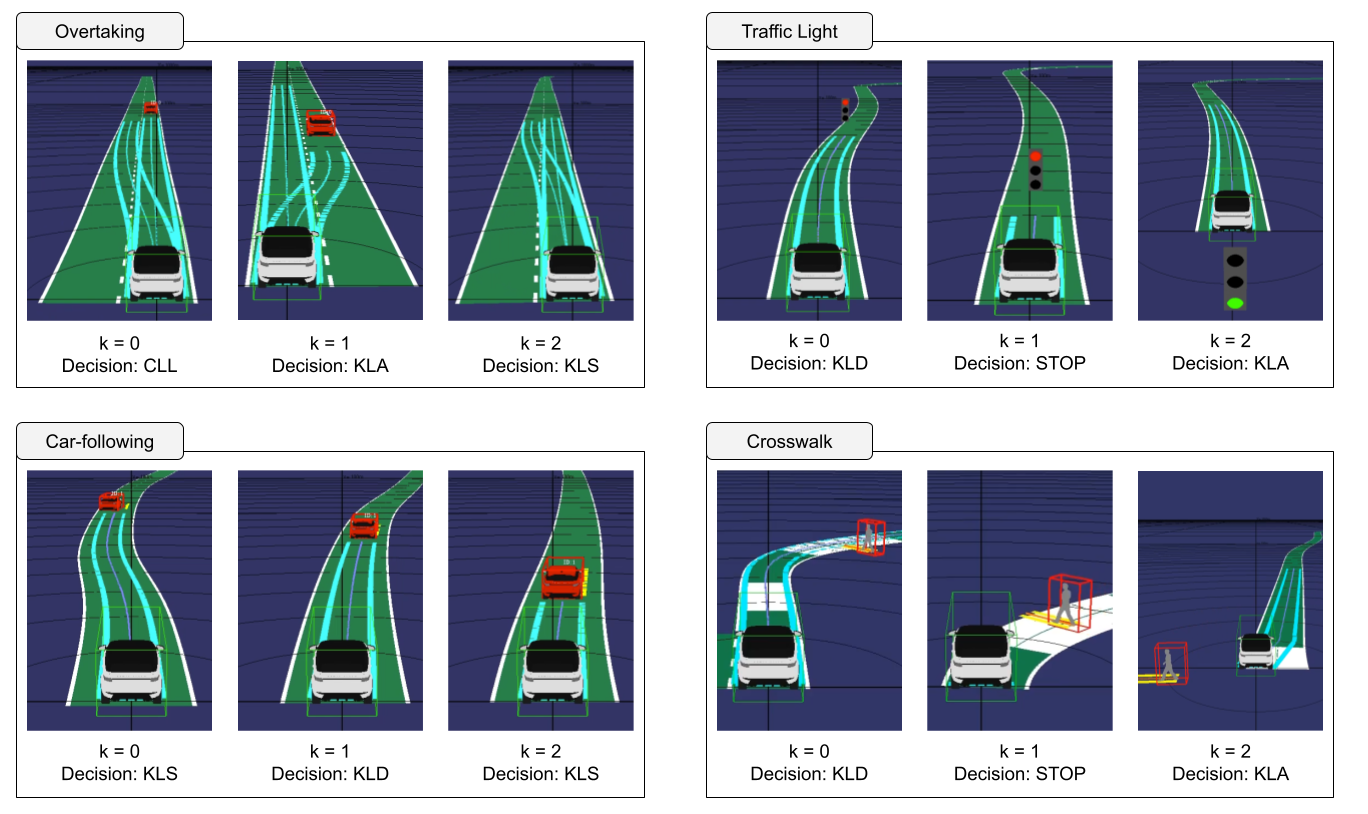}
\caption{Snapshots at different time step k of four different test cases obtained in simulation while testing COR-MP. CLL: Change Lane Left, KLA: Keep Lane Accelerate, KLS: Keep Lane Same speed, KLD: Keep Lane Decelerate.}
\label{fig:simulationResults}
\end{figure*}

To validate our approach, we use RTMaps middleware. Diverse snapshots of some results obtained in simulation are presented in Figure \ref{fig:simulationResults}. The primary objective of these simulation tests is twofold: firstly, to prove the suitability of the Conservation of Resources theory for a decision-making algorithm in automated driving; and secondly, to prove the effectiveness of COR-MP in diverse driving scenarios, including interactions with other road users.
To demonstrate the effectiveness of COR-MP, we evaluate it in the same manner as humans do while learning to drive. Just as a driving exam assesses a person based on their compliance with traffic rules and on its ability to safely share the road with other road users, we evaluate COR-MP similarly. Specifically, we check whether the final decisions proposed by our approach avoid collisions and comply with traffic rules.

Figure \ref{fig:simulationResults} presents four of the test cases used to validate COR-MP. These results were obtained in a closed-loop configuration, where the output of our approach directly affects the lateral and longitudinal movements of the vehicle. Specifically, COR-MP determines both what decisions to make and when to make them.

In these illustrations, the ego-vehicle is represented by the white car, while the interacting vehicles are depicted in red. The trajectories evaluated as feasible for the EV are in cyan, and the predicted trajectories for other interacting objects are depicted in yellow. 

The use case presented on the top left of the illustration is an overtaking scenario on a two-lane road, where an interacting object is static in the middle of the right-most lane. The objective was to determine the decision proposed by COR-MP for the ego-vehicle. As highlighted by the snapshots, our approach evaluates that, in this situation, the best maneuver for the ego-vehicle is overtaking. This result is encouraging, as it aligns with what most drivers would have done.

The one presented at the top right corner illustrates COR-MP's capability to respect traffic rules, including traffic light regulation. This figure shows that the ego stops at a red light and starts to move again once the light turns green. Furthermore, across all the test cases, COR-MP did not propose any maneuvers that would exceed speed limits, cross continuous lanes, or violate any other traffic rules.

The two scenarios at the bottom of Figure \ref{fig:simulationResults} demonstrate the safety aspect of our approach. In the use case depicted in the bottom left figure, the interacting vehicle is driving at a slower speed compared to the speed limit. Our approach ensures safety by adjusting the speed of the ego-vehicle to match with the speed of the interacting vehicle while maintaining a safe longitudinal distance between the EV and the vehicle in front.

The test case presented at the bottom right of Figure \ref{fig:simulationResults} shows that our approach handles and guarantees safety with other road users and particularly with pedestrians. As highlighted, the ego-vehicle stopped just before the crosswalk to allow the pedestrian to pass through and start to move again once the pedestrian had completely crossed the crosswalk.

\subsection{Vehicle Results}
   
\begin{figure}[h]
\centering
\includegraphics[width=0.5\textwidth]{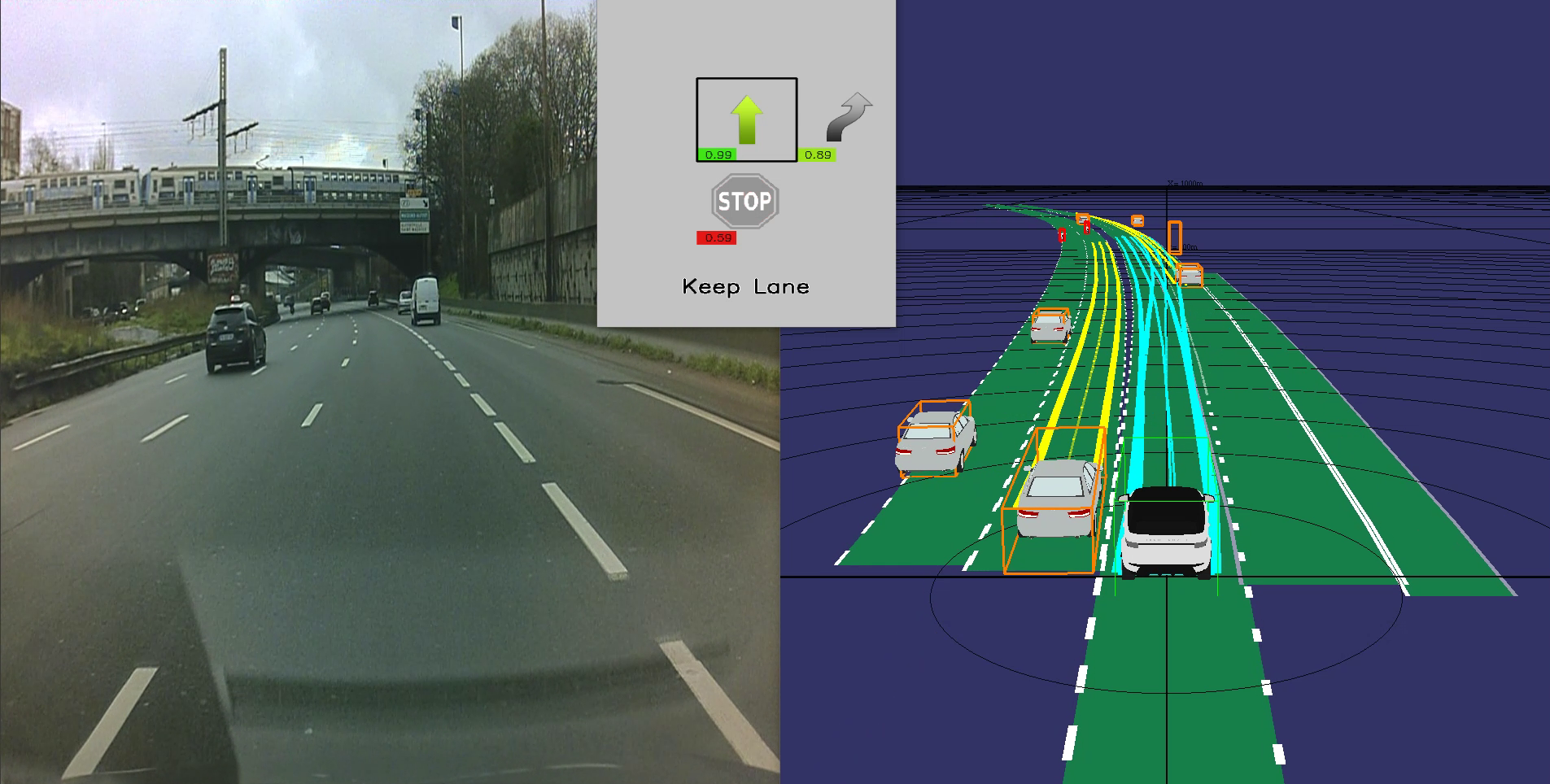}
\caption{Snapshot of the interface feedback while testing COR-MP in a real vehicle on highway. \textit{Left image}: Front camera of the vehicle. \textit{Middle image}: Maneuver viewer showing the decision proposed by our approach. \textit{Right image}: Graphic view of the environment.}
\label{fig:interface}
\end{figure}

\begin{figure}[h]
\includegraphics[width=.5\textwidth]{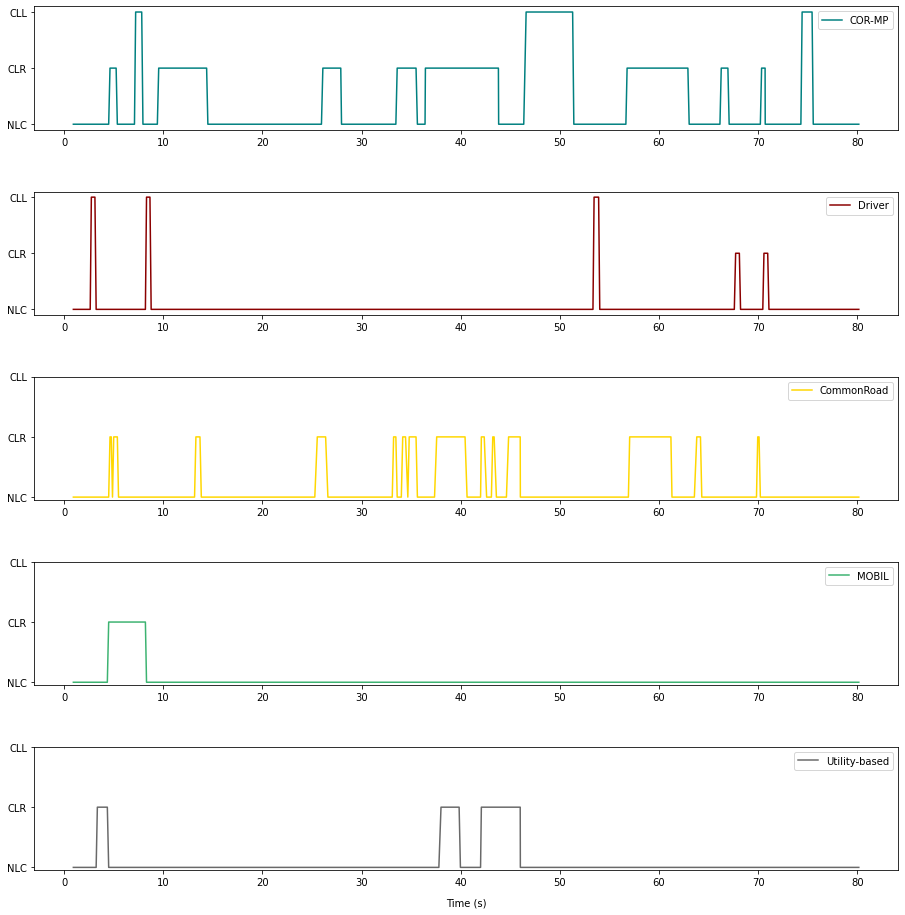} 
\caption{Comparison of lateral decisions between our approach, driver decisions, and diverse baselines on a section of a 5-lane highway. NLC: No Lane Change, CLR: Change Lane Right, CLL: Change Lane Left.}
\label{fig:VehicleResults}
\end{figure}
     
\begin{figure}[h]
\centering
\includegraphics[width=0.4\textwidth]{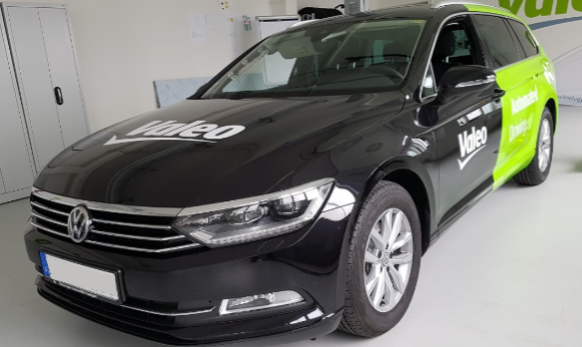}
\caption{The vehicle serves to test our approach.}
\label{fig:cruise4u}
\end{figure}

COR-MP has also been tested on a real vehicle in highway and urban environments. It was experimented with in an open-loop configuration, meaning that the output of our approach did not directly affect the ego-vehicle movement. Here, the goals were twofold: first, to analyze how our approach will adapt to real data which are different from simulation as they contain uncertainties; second to determine if COR-MP would behave safely. Figure \ref{fig:cruise4u} shows the vehicle used for testing and validating our approach. Figure \ref{fig:interface} presents a snapshot of COR-MP running in real-time with data from the real vehicle.

Determining the effectiveness of a decision-making algorithm for automated driving is complex. This complexity arises because, for the same situation, the adopted decision may vary depending on individuals and their personal driving preferences. That is why, initially, we evaluated our approach by analyzing whether it produces collision-free decisions and adheres to traffic rules.

Subsequently, the effectiveness was evaluated by comparing the decisions made by our approach with those made by the driver. Since one of the driving modes of COR-MP is designed to align with the decisions a regular human would make. We instructed the pilot to drive as regularly and rationally as possible without any information about the proposed outputs of COR-MP, so as not to influence his decisions. 

Figure \ref{fig:VehicleResults} compares the lateral decisions proposed by our approach, the driver, and some from literature baselines. 
Since we predict a trajectory for each maneuver that closely represents the one the ego vehicle will follow, we can evaluate each maneuver using the function \textit{SM1} from the CommonRoad benchmark \cite{CommonRoad}. Decisions proposed by this benchmark are depicted in yellow in Figure \ref{fig:VehicleResults}. The graph in green represents lateral decisions from the Minimizing Overall Braking Induced by Lane Changes (MOBIL) model \cite{MOBIL}. Finally, the graph at the bottom of Figure \ref{fig:VehicleResults} represents lateral decisions proposed by a standard utility-based maneuver planner that optimizes safety, a-priori lane, time, and comfort parameters.

These results have been obtained on a portion of a 5-lane highway, as shown by the front camera of the vehicle in Figure \ref{fig:interface}. At time t = 0 second, the ego-vehicle was located in the right-most lane of the road. 

\begin{table}[!t]
\centering
\begin{tabular}{lccc}
   \textbf{Method} & \textbf{Average speed \(\Bar{v}\) (m/s)} & \textbf{Average acceleration \(\Bar{a}\) (m/s²)}\\
    \hline
    CommonRoad&24.058&0.294\\ 
    Driver&22.589&0.017\\ 
    \textbf{COR-MP}&22.155&-0.087\\
    \hline
\end{tabular}
\caption{Efficiency and comfort performance results.}
\label{tab:efficiency}
\end{table}

The first notable observation we noticed is that, during this portion of highway, only COR-MP proposed changing to the left lane. In fact, when the ego vehicle was getting closer to another vehicle, COR-MP chose to change lane instead of decelerating, whereas other baseline methods opted for deceleration in the same lane.
Another interesting point is that each time COR-MP proposed a change to the left lane, the driver decided to change to the left lane a few seconds later. Furthermore, near the end of the graph, we noticed that COR-MP proposed a lane change to the left lane while the driver did not change lanes. This difference arises because the driver aimed to stay in the exit lane to follow its route, whereas COR-MP lacked route information, which explains the mismatch. 

The main differences observed between COR-MP and the driver relate to right lane changes. For this maneuver, our approach can be considered more altruistic. Because, multiple times, when there is a right lane available and it is evaluated as safe to reach it, COR-MP proposes to change to the right lane to make the overtaking lane free for the other drivers -  a decision not constantly made by each human, as the driver demonstrates. 

During this section of the 5-lane highway, both COR-MP and the driver decide to overtake another vehicle twice, and approximately at the same time. This is encouraging for validating the effectiveness of our approach, as one of the goals of COR-MP is to behave in a human-like manner.

The vehicle test lasted about 45 minutes and included driving in both urban and highway environments. To evaluate longitudinal decisions, we assessed both efficiency and comfort aspects by analyzing speed and acceleration. Table \ref{tab:efficiency} presents the average of these values proposed by our approach, the driver, and from the CommonRoad baseline \cite{CommonRoad}, using the function \textit{SM1}.

During the urban portion of the trip, the EV interacted with a high density of other road users. Additionally, the driver did not always maintain a safe longitudinal distance from the vehicle ahead, which contributed to the negative average acceleration value obtained by our approach. Since COR-MP was set to Regular driving mode, meaning the resource \textit{Safety} was prioritized over the others, most of the decisions proposed by COR-MP maintained safe longitudinal and lateral distances with other road users. 

Finally, throughout the entire vehicle journey, we observed that our method never proposed a decision that could lead to a collision. The resource \textit{Safety} remained acquired during the whole trip, guaranteeing the respect of the safety aspect continuously. Moreover, each decision conformed to traffic rules, including speed limits, avoidance of crossing continuous lines, and not entering restricted lanes




\section{Conclusions and Future Work}
\label{section:Conclu&FutureWork}
In this study, we introduced COR-MP, a new model for decision-making in automated driving. This model is based on the Conservation of Resources theory, a psychological concept used to understand human behavior in stressful situations. We classify our model as a psychological model and categorize it as a sub-model of utility-based techniques in decision-making. 

Our approach has been tested in simulation and in a real vehicle. Based on the results, COR-MP has proven that the COR theory is suitable for maneuver planning in automated driving. However, additional extended tests are planned to further validate and enhance our approach.
Evaluating a decision and determining its effectiveness is complex, as viewpoints may vary depending on individuals. However, in automated driving, it is universally agreed that the paramount parameter to respect while driving is safety. That is why, the effectiveness of our approach has been demonstrated by the fact that no decision proposed by COR-MP led to a collision or a non-respect of the traffic rules. Additionally, we noticed that some decisions made by our approach are similar to those of a regular driver. This is encouraging, as we aimed to develop a human-like maneuver planner.

Interpretability is a crucial point, as a decision-making algorithm should explain why a decision has been taken or not. COR-MP ensures a certain level of interpretability, as the logic behind each decision can be discerned by analyzing resources' value and state. 



While driving in the real car, uncertainties were present, notably from the perception stage. This may lead to inappropriate decisions. In future work, we aim to take into consideration uncertainty to enhance the effectiveness of COR-MP in various driving situations. 
We also want to extend COR-MP to evaluate a sequence of maneuvers instead of only one maneuver. This would result in a more human-like decision while driving. 

\section*{Acknowledgment}
The authors would like to express their gratitude to Nelson de Moura and Tiago Rocha Gonçalves for their feedback concerning this study.

\bibliographystyle{IEEEtran}
\bibliography{bibliography}

\end{document}